# Role of Ontology in Semantic Web Development

Zeeshan Ahmed and Detlef Gerhard

Mechanical Engineering Information and Virtual Product Development (MIVP),
Vienna University of Technology,
1060 Getreidemarkt 9/307 Vienna, Austria
{zeeshan.ahmed, detlef.gerhard} @tuwien.ac.at

World Wide Web (WWW) is the most popular global information sharing and communication system consisting of three standards .i.e., Uniform Resource Identifier (URL), Hypertext Transfer Protocol (HTTP) and Hypertext Mark-up Language (HTML). Information is provided in text, image, audio and video formats over the web by using HTML which is considered to be unconventional in defining and formalizing the meaning of the context. Most of the available information is unstructured and due to this reason it is very difficult to extract concrete information. Although some search engines and screen scrapers are developed but they are not quite efficient and requires excessive manual preprocessing e.g. designing a schema, cleaning raw data, manually classifying documents into taxonomy and manual post processing. To increase the integration and interoperability over the web the concept of "Web Service" was introduced. Due to the dynamic nature Web Service become very popular in industry in short time but with the passage of time due to the heavily increase in number of services the problems of end-to-end service authentication, authorization, data integrity and confidentiality were identified [1].To cope with the existing web based problems .i.e., Information filtration, security, confidentiality and augmentation of meaningful contents in mark-up presentation over the web a semantic based solution "Semantic Web" was introduced by Tim Berners Lee [5]. Semantic Web is an intelligent incarnation and advancement in World Wide Web to collect, manipulate and annotate the information by providing categorization, uniform access to resources and structuring the information in machine process able format. To structure the information in machine process able semantic models Semantic Web have introduced the concept of "Ontology" [2].

Ontology is the collection of interrelated semantic based modeled concepts based on already defined finite sets of terms and concepts used in information integration and knowledge management. To obtain the desired results Ontology is categorized in to three categories .i.e., Natural Language Ontology (NLO), Domain Ontology (DO) and Ontology Instance (OI). NLO creates relationships between generated lexical tokens obtained from natural language statements, DO contains the knowledge of a particular domain and OL generates automatic object based web pages.

Ontology construction is a highly relevant research issue depending on the extraction of information from web and emergence of ontologies. Ontologies are constructed using some ontology supporting languages like RDF, OWL etc. and connected to each other in a decentralized manner to clearly express semantic contents and arrange semantic boundaries to extract concrete information. Ontology is heavily contributing in industry by supporting the development of advanced language processing tools and large linguistics resources, some in time proposed and developed ontology based solutions are Semantic Desktop [3], Semantic Security Web Services (SSWS) [1] and Cultural Heritage and the Semantic Web [4].

Where Ontology is helping in providing solutions to the document identification, end-to-end service authentication, authorization, data integrity, confidentiality, organizing and sharing of isolated pieces of information problems there it is also facing some limitations which are

- Natural language parsers can only work over a single statement at a time.
- Quite impossible to define the boundaries of ontologies of a particular domain's abstract model.
- Automatic ontology creations, automatic emergence of ontologies to create new ontologies and identification of possible existing relationships between classes to draw the taxonomy hierarchy automatically is required.
- Ontology validators are restricted and not capable of validating all kind of ontologies e.g. based on complex inheritance relationship.
- Domain specific ontologies are highly dependent on the domain of the application and because of this dependency it is not possible to find out the general purpose ontologies from them.
- The process of semantic enrichment reengineering for the web development consists of relational metadata, required to be developed at high speed and in low cost depending on proliferation of ontologies, which is currently also not possible.

Due to these limitations in ontology Semantic Web is not currently succeeded in attaining the actual goals of completely structuring the information over the web in machine process able format and making advanced knowledge modeled system.

## References

1. Grit, Dr, Son, N, & Andrew, T, 2004, "OWL-S Semantics of Security Web Services: a Case Study", pp. 240–253, Springer-Verlag J. Davies et al. (Eds.): ESWS, LNCS 3053, Berlin Heidelberg Germany
2. Heiner, S 2002, "Approximate Information Filtering on the Semantic Web", pp. 114–128, Springer-Verlag M. Jarke et al. (Eds.): KI, LNAI 2479, Berlin Heidelberg Germany
3. Mark, S, Pierre, S, Leo, S & Andreas, D, 2006, "Increasing Search Quality with the Semantic Desktop in Proposal Development", In the proceedings of Practical Aspects of Knowledge Management 6th International Conference. (PAKM), Vienna Austria
4. V.R, Benjamins, J. Contreras, M. Blázquez, J.M. Dodero, A. Garcia, E. Navas, F. Hernandez & C. Wert, 2004, "Cultural Heritage and the Semantic Web", pp. 433-444, Springer-Verlag J. Davies et al. (Eds.): ESWS LNCS 3053, Heidelberg Germany
5. Tim, B. Lee, James, H & Ora, L, 2001, The Semantic Web, A new form of Web content that is meaningful to computers will unleash a revolution of new possibilities, Retrieved June 30, 2007 from http://www.geodise.org/useful_links/link_semantic.htm